\documentclass[10pt,twocolumn,letterpaper]{article}

\usepackage{iccv}
\usepackage{times}
\usepackage{epsfig}
\usepackage{graphicx}
\usepackage{amsmath}
\usepackage{amssymb}
\usepackage{booktabs} 
\usepackage{algorithm}
\usepackage{algorithmic}
\usepackage{amsthm}
\usepackage{diagbox}
\usepackage{float}
\usepackage{epstopdf}
\usepackage{enumitem}
\usepackage{multirow}
\usepackage{marvosym}
\usepackage{bbm}
\usepackage{colortbl}
\usepackage{makecell}

\usepackage[T1]{fontenc}
\usepackage[latin9]{inputenc}
\usepackage{array}

\usepackage[ruled,vlined,algo2e]{algorithm2e}
\usepackage[dvipsnames]{xcolor}

\definecolor{citecolor}{RGB}{119,185,0} 
\usepackage[pagebackref=true,breaklinks=true,letterpaper=true,colorlinks,citecolor=citecolor,bookmarks=false]{hyperref}

\usepackage[capitalize]{cleveref}
\crefname{section}{Sec.}{Secs.}
\Crefname{section}{Section}{Sections}
\Crefname{table}{Table}{Tables}
\crefname{table}{Tab.}{Tabs.}

\DeclareMathOperator*{\argmax}{arg\,max}

\usepackage{color,colortbl}
\definecolor{ballblue}{rgb}{0.13, 0.67, 0.80}
\definecolor{ggray}{rgb}{0.92, 0.92, 0.99}
\definecolor{amaranth}{rgb}{0.90, 0.17, 0.31}
\definecolor{olive}{rgb}{0.5, 0.5, 0.0}
\definecolor{lightblue}{RGB}{72, 187, 231}
\newcolumntype{a}{>{\columncolor{lightblue!15}}c}
\definecolor{blush}{rgb}{0.87, 0.36, 0.51}
\newcommand{\ar}{\texttt{ART}}

\newcommand*\samethanks[1][\value{footnote}]{\footnotemark[#1]}

\iccvfinalcopy 

\def\iccvPaperID{1355} 

\ificcvfinal\fi

\begin{document}

\title{Activate and Reject: Towards Safe Domain Generalization under Category Shift}

\author{Chaoqi Chen$^{1}\thanks{First two authors contributed equally.}$, Luyao Tang$^{2}\samethanks$, Leitian Tao$^{3}$, Hong-Yu Zhou$^1$, Yue Huang$^2$, Xiaoguang Han$^4$\thanks{Corresponding authors.}, Yizhou Yu$^{1\dag}$\\
	{$^1$~The University of Hong Kong}\quad{$^2$~{Xiamen University}}\\
	{$^3$}~{University of Wisconsin - Madison}\quad{$^4$~The Chinese University of Hong Kong (Shenzhen)}\\
	{\tt\small cqchen1994@gmail.com, lytang@stu.xmu.edu.cn, taoleitian@gmail.com, whuzhouhongyu@gmail.com}\\   
	{\tt\small yhuang2010@xmu.edu.cn, hanxiaoguang@cuhk.edu.cn, yizhouy@acm.org}
}


\maketitle
\ificcvfinal\thispagestyle{empty}\fi

\begin{abstract}
Albeit the notable performance on in-domain test points, it is non-trivial for deep neural networks to attain satisfactory accuracy when deploying in the open world, where novel domains and object classes often occur. In this paper, we study a practical problem of Domain Generalization under Category Shift~(DGCS), which aims to simultaneously detect unknown-class samples and classify known-class samples in the target domains. Compared to prior DG works, we face two new challenges: 1) how to learn the concept of ``unknown'' during training with only source known-class samples, and 2) how to adapt the source-trained model to unseen environments for safe model deployment. To this end, we propose a novel \emph{Activate and Reject (\ar)} framework to reshape the model's decision boundary to accommodate unknown classes and conduct post hoc modification to further discriminate known and unknown classes using unlabeled test data. Specifically, during training, we promote the response to the unknown by optimizing the unknown probability and then smoothing the overall output to mitigate the overconfidence issue. At test time, we introduce a step-wise online adaptation method that predicts the label by virtue of the cross-domain nearest neighbor and class prototype information without updating the network's parameters or using threshold-based mechanisms. Experiments reveal that \ar\ consistently improves the generalization capability of deep networks on different vision tasks. For image classification, \ar\ improves the H-score by 6.1\% on average compared to the previous best method. For object detection and semantic segmentation, we establish new benchmarks and achieve competitive performance.

\end{abstract}

\vspace{-0.5cm}
\section{Introduction}

Deep neural networks have achieved unprecedented success in a myriad of vision tasks over the past decade. 
Despite the promise, a well-trained model deployed in the open and ever-changing world often struggles to deal with the domain shifts---the training and testing data do not follow the independent and identically distributed (i.i.d) assumption, and therefore deteriorates its safety and reliability in many safety-critical applications, such as autonomous driving and computer-aided disease diagnosis.
This gives rise to the importance of Domain Generalization~(DG)~\cite{zhou2022domain,wang2022generalizing}, \emph{a.k.a.} out-of-distribution~(OOD) generalization, 
which aims at generalizing predictive models trained on multiple (or a single) source domains to unseen target distributions. 

\begin{figure}[!t]
\centering	\includegraphics[width=0.46\textwidth]{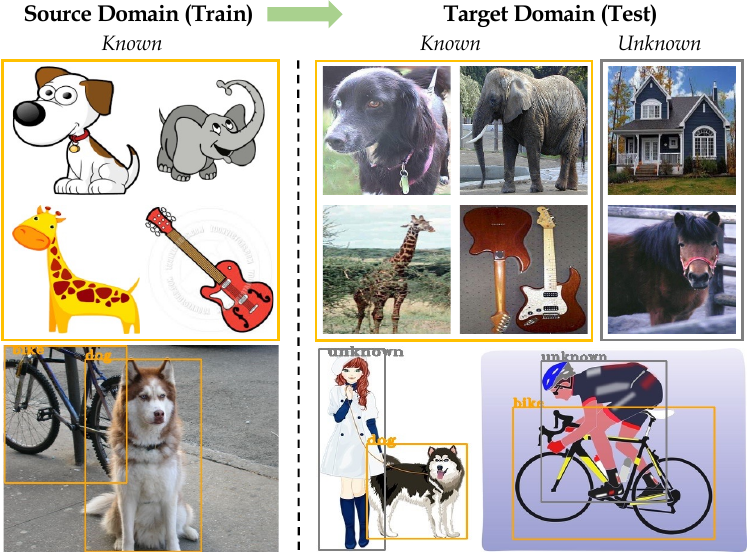}
\caption{DGCS in image classification and object detection tasks.}
\label{fig:illustration}
\vspace{-0.5cm}
\end{figure}

In order to unearth domain-agnostic knowledge and alleviate domain-specific components,
a plethora of DG algorithms have been proposed, spanning invariant risk minimization~\cite{arjovsky2019invariant,ahuja2020invariant}, augmentation~\cite{volpi2018generalizing,xu2021robust,zhou2021domain,chen2022mix}, feature disentanglement~\cite{piratla2020efficient,liu2021learning,zhang2022towards}, meta-learning~\cite{li2018learning,li2019episodic,dou2019domain}, to name a few.  
Among them, a common assumption is that the label spaces of source and target domains are identical, which may not always hold in practice.
Suppose that we wished to deploy modern vision systems to recognize objects in an autonomous vehicle.
When only the environment (\emph{e.g.,} weather and illumination) and appearance (\emph{e.g.,} size and viewpoint) of previously seen objects can change,
principled approaches are capable of correcting for the potential shifts on the fly.
But what if the sudden arrival of new objects in an ever-changing world? 
Most existing DG methods will break and may even result in catastrophe, raising strong concerns about model reliability.
Although several prior arts~\cite{shu2021open,zhu2022crossmatch} have explored the open DG scenarios, 
the ``adaptivity gap''~\cite{dubey2021adaptive} between training and test distributions still hinders safe deployment of source-learned models~\cite{hendrycks2021unsolved}. 





To this premise, we challenge the status quo by raising an open question:
\emph{can deep models learn what they don't know during training and subsequently adapt to novel environments at test-time for safe model deployment?}
Thus, we consider a more realistic scenario namely Domain Generalization under Category Shift~(DGCS) (see Fig.~\ref{fig:illustration}), 
wherein the source-trained model is expected to simultaneously detect unknown-class samples and categorize known-class samples under the presence of domain shifts.
The core challenges are: \emph{(i)} no unknown-class data is available in training and \emph{(ii)} the mixture of domain and label shifts during test time.  
In this paper, we present a simple yet effective framework---\textbf{A}ctivate and \textbf{R}ejec\textbf{T}~(dubbed \ar), 
which reshapes the model's decision boundary to accommodate unknown classes and adjusts the final prediction to reconcile the intrinsic tension between domain and label shifts.
\ar\ encapsulates two key components: \emph{(i)} Unknown-aware Gradient Diffusion~(UGD) to make the classifier give response to unknown dimension and smooth the decision boundary to mitigate overconfidence;
\emph{(ii)} Test-time Unknown Rejection~(TUR) to conduct \emph{post hoc} modification to the learned classifier's final predictions, 
making the decision boundaries of different classes closer to the well-behaved case.

Specifically, the logit of unknown class is activated by minimizing the negative log-likelihood regarding unknown probability.
However, we find that the learned probability will be suppressed due to the overconfidence \emph{w.r.t.} known classes.
Thus, we introduce a smoothed cross-entropy loss to promote the response to the unknown by adding the penalty on the $L_2$ norm of the logits and using a temperature scaling parameter, where the former mitigates the excessive increase of the logit norm while the latter magnifies the effect of logit penalty.
Due to the unavailability of real target data in training, 
the source-trained decision boundaries between known and unknown classes may still be ambiguous.
Therefore, TUR refines the source-trained classifier using unlabeled test data in an online adaptation manner.
To be specific, 
TUR first determines if the input belongs to known classes or not via a cross-domain nearest neighbor search, based on prototype information and cyclic consistent constraint;
otherwise, the prediction will be made by a parallel module that measures the input's similarity with a set of dynamically-updated target prototypes.  
TUR is training-free (no backward passes) and does not rely on threshold-based criteria nor impose any distributional assumptions.

Our key contributions are summarized as follows:
\vspace{-2mm}
\begin{itemize}
 \item We study a challenging DG problem~(DGCS) and  propose a principled framework (\ar) to jointly consider domain shift, label shift, and adaptivity gap. 
\vspace{-2mm}
 \item We propose an unknown-aware training objective to activate the unknown's logit and alleviate the overconfidence issue, and an online adaptation strategy to perform post hoc modification to the learned classifier's prediction at test-time without additional tuning.
\vspace{-2mm}
\item Extensive experiments show that \ar\ achieves superior performance on a wide range of tasks including image classification, object detection, and semantic segmentation. In particular, on four image classification benchmarks (PACS, Office-Home, Office-31, and Digits), \ar\ improves the H-score by 6.1\% on average compared to the previous best method.

\end{itemize}

	


\section{Related Works}

\begin{figure*}[t]
	\centering
	\small
	\setlength\tabcolsep{1mm}
	\renewcommand\arraystretch{0.1}
	\begin{tabular}{cccc}
		\includegraphics[width=0.24\linewidth]{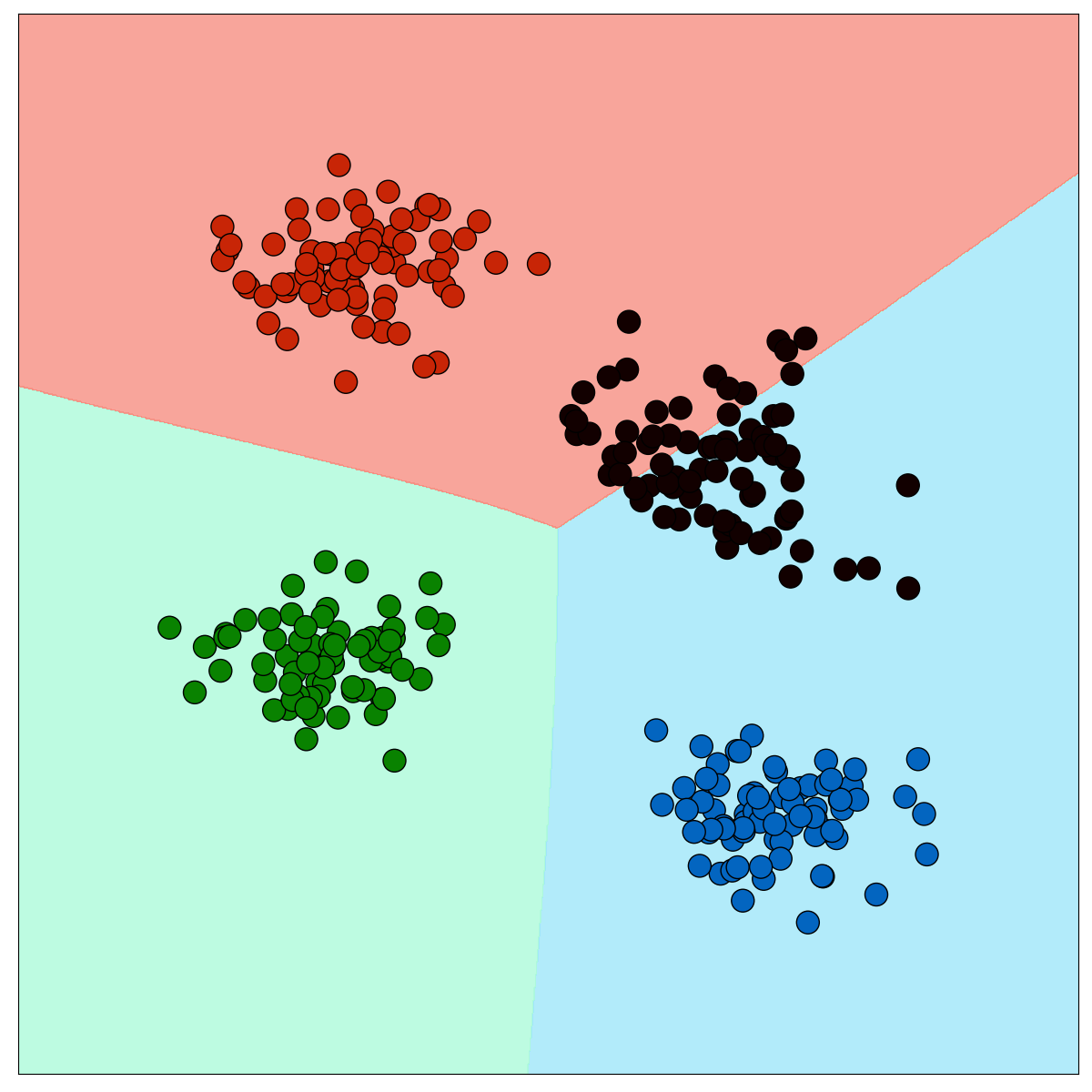} &
		\includegraphics[width=0.24\linewidth]{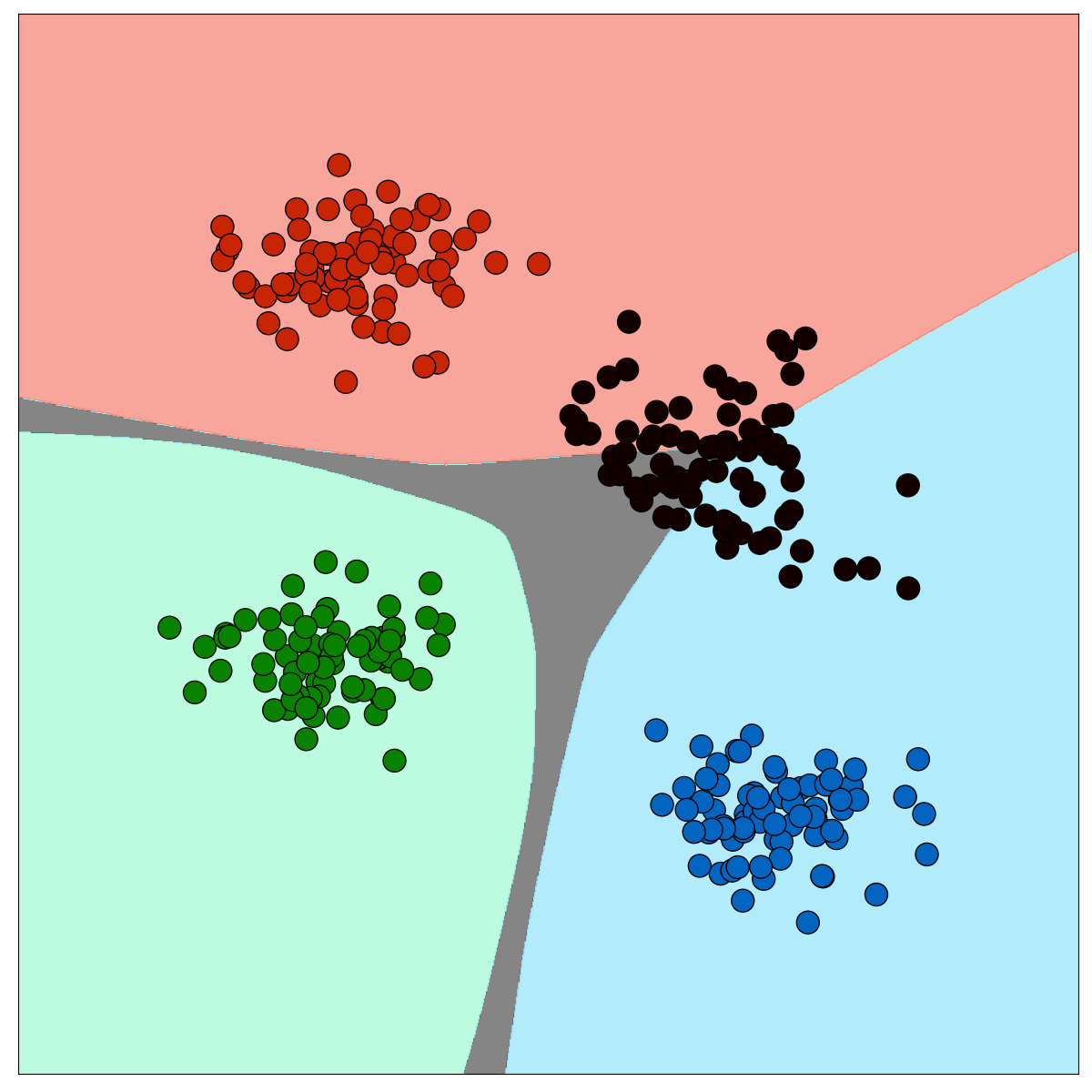} & 
		\includegraphics[width=0.24\linewidth]{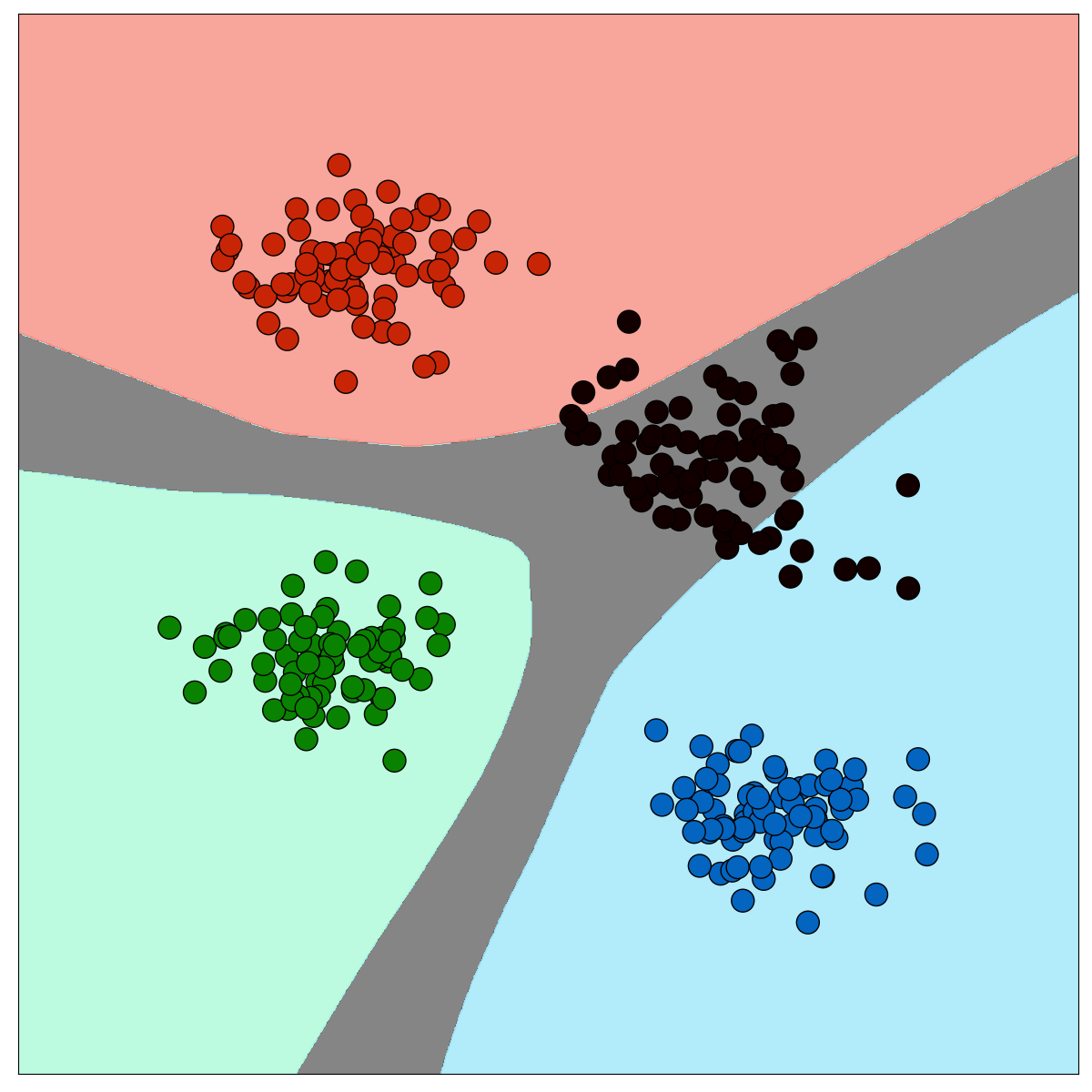} &
		\includegraphics[width=0.24\linewidth]{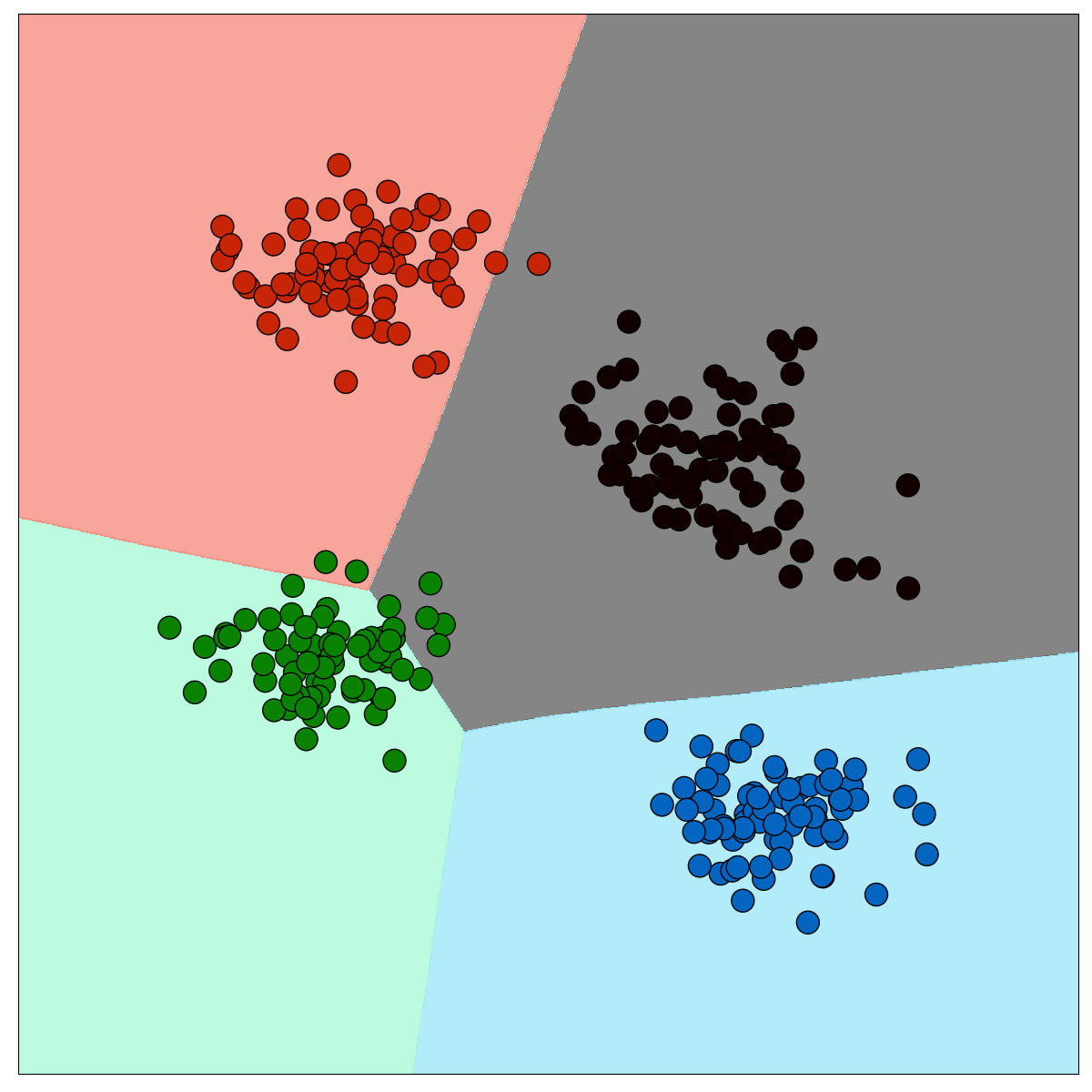} \\
		(a)  & (b)  & (c)  & (d) \\
	\end{tabular}
        \vspace{1mm}
	\caption{Toy example illustrating the decision boundaries learned by different methods. 
		We generate isotropic Gaussian blobs with 4 classes.
		Red, green, and blue points indicate the known-class samples. Black points  denote unknown-class samples, which \emph{are unavailable during training.}
		\textbf{(a)} Train with standard CE loss, \emph{i.e.,} vanilla ($|\mathcal{C}_s|$+1)-way classifier in DGCS. 
            \textbf{(b)} Train with our unknown activation loss $\mathcal{L}_{\text{UA}}$.
		\textbf{(c)} Train with full UGD loss $\mathcal{L}_{\text{UGD}}$.		
		\textbf{(d)} The result of \ar\ (UGD + TUR). 
            This figure is best seen in color.	
	}
	\label{fig:toy_example}
\end{figure*} 

\paragraph{Domain Generalization~(DG).}
The objective of DG is to learn representations that are independent of domain-specific factors and thus can extrapolate well to unseen test distributions.   
This is typically achieved by invariant learning and robust learning. 
Current approaches can be broadly categorized into feature matching~\cite{li2018deep,dg_mmld,zhu2022localized,chen2023domain}, decomposition~\cite{rojas2018invariant,piratla2020efficient,christiansen2021causal,mahajan2021domain,sun2021recovering,liu2021learning,zhang2022towards,zhou2023transformer}, augmentation~\cite{volpi2019addressing,zhou2020deep,zhou2020learning,xu2021robust,nam2021reducing,zhou2021domain,xu2021fourier,zhang2022exact,chen2022self,chen2022compound}, and meta-learning-based~\cite{li2018learning,li2019feature,li2019episodic,dou2019domain,chen2022ost} approaches. 
To adapt to complex real-world applications, very recently, several works~\cite{shu2021open,zhu2022crossmatch,yang2022one} consider the existence of both known and unknown classes in new DG settings, such as open DG~\cite{shu2021open} open-set DG~(OSDG)~\cite{zhu2022crossmatch}. 
Shu~\etal~\cite{shu2021open} assume that both source and target domains have different label spaces and introduce novel augmentation strategies to augment domains on both feature- and label-level. 
Zhu~\etal~\cite{zhu2022crossmatch} generate auxiliary samples via an adversarial data augmentation strategy~\cite{volpi2018generalizing} and enhance unknown class identification with multi-binary classifiers. 
Yang~\etal~\cite{yang2022one} introduce an additional CE loss based on the assumption that any non-ground-truth category can be viewed as unknown categories. 
However, these works rely on additional training modules and heuristic thresholding mechanism~\cite{zhu2022crossmatch} or impose a strong distributional assumption of the feature space regarding known and unknown data~\cite{yang2022one}. 
In addition, Dubey~\emph{et al.}~\cite{dubey2021adaptive} reveal that there will always be an ``adaptivity gap'' when applying the source-learn model to target domains without further adaptation. 
How to endow the source model with the capability of identifying unseen open classes and safely adapting the learned classifier to unlabeled test samples is yet to be thoroughly studied. 

\paragraph{Domain Adaptation~(DA).}
DA~\cite{pan2011domain,long2015learning,ganin2015unsupervised,chen2019progressive,chen2022relation} aims to improve the performance of the learned model on the target domain using labeled source data and unlabeled target data. In addition to the close-set setting, many new and practical DA paradigms have been proposed, such as partial~\cite{zhang2018importance,cao2019learning}, open-set~\cite{panareda2017open,saito2018open,kundu2020towards,bucci2020effectiveness,chen2021dual,liu2022unknown}, universal~\cite{you2019universal,saito2020universal}, and source-free~\cite{xia2021adaptive,yang2021exploiting,ding2022source,zhang2022divide,yang2022attracting}. In particular, open-set DA~(OSDA) and source-free DA~(SFDA) are closely related to the problem explored in this paper. 

\paragraph{Test-Time Adaptation~(TTA).}
For DG, due to the inaccessibility of target data during training, it is natural to solve the adaptivity gap~\cite{dubey2021adaptive} with TTA strategies.
Adaptive methods~\cite{liang2020we,sun2020test,wang2021tent,iwasawa2021test,pandey2021generalization,chen2022contrastive,zhang2022memo,chen2023improved} have been proposed to refine the matching process between target test data and source-trained models in an online manner, \emph{i.e.,}  all test data can be accessed only once. 
Tent~\cite{wang2021tent} proposes to reduce the entropy of model's predictions on test data via entropy minimization.
T3A~\cite{iwasawa2021test} introduces a training-free approach by classifying each test sample based on its distance to a dynamically-updated support set.
Despite the promising results on closed-set classes, these approaches fail to deal with open-set samples and thus lead to semantic mismatching.

\paragraph{Out-of-Distribution Detection~(OD).}
A separate line of work studies the problem of OD~\cite{yang2021generalized,tao2023non}, 
which aims to identify novel examples that the network has not been exposed to at the training phase. 
Mainstream OD methods are devoted to design OOD scoring functions, \emph{e.g.,} confidence-based approaches~\cite{bendale2016towards,hendrycks2017baseline,huang2021mos}, distance-based score~\cite{lee2018simple,2021ssd,sun2022knnood}, and energy-based score~\cite{liu2020energy,sun2021react}.  
The main difference between OD and our problem is that the former is a binary classification problem and does not account for the domain and label shifts between training and test data at the same time. 

\begin{table}[t]
\small
\centering
\caption{
Comparison of different problem settings.
$(X_s, Y_s)$ and $X_t$ are the labeled source and unlabeled target data respectively. Fine-tune means to update the model's parameters. Adjustment means making post-hoc modifications to the model's predictions. 
}
\label{tab:setting}
\vspace{3mm}
\scalebox{0.7}{
\begin{tabular}{lccccc}
\toprule
\multirow{2}{*}{\textbf{Problem Setting}} & \textbf{Training} & \multicolumn{4}{c}{\textbf{Test-time}} \\
\cmidrule(lr){2-2} 
\cmidrule(lr){3-6}
 & Data & Domain Shift & Open Class & Fine-tune & Adjustment   \\ 
\midrule
OD~\cite{hendrycks2017baseline,sun2022dice} & $X_s, Y_s$ & $\times$ & \checkmark & $\times$ & $\times$ \\
OSDA~\cite{saito2018open,bucci2020effectiveness} & $X_s, Y_s, X_t$ & \checkmark & \checkmark & $\times$ & $\times$ \\
SFDA~\cite{you2019universal,zhang2022divide} & $X_s, Y_s, X_t$ & \checkmark & $\times$ & \checkmark & $\times$ \\
TTA~\cite{sun2020test,wang2021tent,zhang2022memo} & $X_s, Y_s$ &  \checkmark & $\times$ & \checkmark & \checkmark \\
OSDG~\cite{zhu2022crossmatch,yang2022one} & $X_s, Y_s$ &  \checkmark & \checkmark & $\times$ & $\times$ \\
\textbf{Ours} & $X_s, Y_s$ & \checkmark & \checkmark & $\times$ & \checkmark \\
\bottomrule
\end{tabular}}
\end{table}

\paragraph{Discussion.}
We provide a comparison of the problem settings among different methods in Tab.~\ref{tab:setting}.
OSDA and SFDA optimize offline with target data and specific learning objectives, while \ar\ only adjusts the classifier in an online manner. 
TTA usually needs to update the trained model's parameters (\emph{e.g.} entropy minimization~\cite{wang2021tent,zhang2022memo}) and a batch of data, while our TUR is fully training-free and can be performed on single test samples.
These promising properties make the proposed approach more suitable for DG. 
Compared to OSDG, our setting allows training-free test-time adjustment for adapting source-trained models to novel environments, largely mitigating the potential adaptivity gap.




\section{Methodology}

\begin{figure*}[t]
	\centering
	\includegraphics[width=0.95\textwidth]{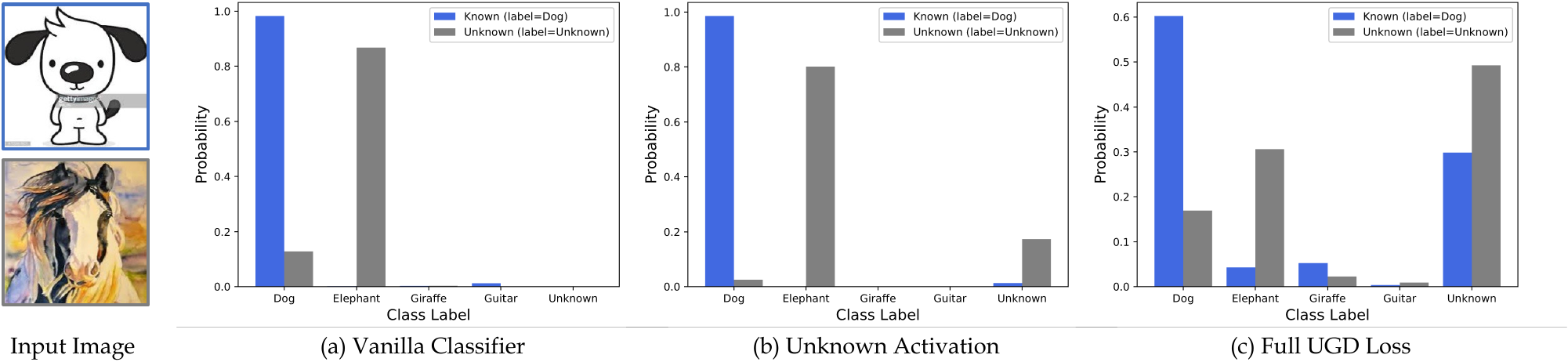}
	\caption{The softmax outputs of different training methods regarding two input images on the PACS~\cite{li2017deeper} benchmark. Source Domain: \emph{Cartoon}, Target Domain: \emph{Art}. 
 Known: \emph{dog} from Domain \emph{Cartoon}, Unknown: \emph{horse} from Domain \emph{Art}.
    }
	\label{fig:prob}
\end{figure*}

\subsection{Preliminary and Motivation}
\label{sec:motivation}

\paragraph{Notation.}
In DGCS, we have a single source domain $\mathcal{D}_s = \{(x_s^i, y_s^i)\}_{i=1}^{n_s}$ of $n_s$ labeled samples and multiple (or a single) unseen target domains $\mathcal{D}_t=\{\mathcal{D}_t^1,...,\mathcal{D}_t^M\}$, where $M \geq 1$ and $\mathcal{D}_t^m = \{(x_t^j, y_t^j)\}_{j=1}^{n_t^m}$.  
$\mathcal{D}_s$ and $\mathcal{D}_t$ are sampled from probability distributions $p_s(x,y)$ and $p_t(x,y)$ respectively.
DGCS jointly considers two distribution shifts: \emph{(i)} class-conditional shift where $p_s(y|x) \neq p_t(y|x)$, and \emph{(ii)} label shift where $p_s(y) \neq p_t(y)$.
Specifically, assume that $\mathcal{C}_{s}$ and $\mathcal{C}_{t}$ are the source and target class sets, respectively.
DGCS dictates $\mathcal{C}_{s} \subset \mathcal{C}_{t}$ and  $\mathcal{C}_t^u = \mathcal{C}_{t}\setminus\mathcal{C}_{s}$ is called \textit{unknown} classes. 
Note we take all unknown classes as a whole even though there can be multiple classes.
The objective of DGCS is to train a model on $\mathcal{D}_s$ to classify all target instances from $\mathcal{D}_t$ into $|\mathcal{C}_{s}|+1$ classes.

\vspace{-0.5cm}
\paragraph{Motivation.}
Before formally introducing technical details, we discuss the motivation of our method using toy data.   
Since the decision boundaries are learned by known classes only,
the unknown target samples tend to lie out of the support of source training data (\emph{i.e.,} low-density regions~\cite{grandvalet2005semi}) and are ambiguous for the decision boundaries. 
On the other hand, 
as shown in Fig.~\ref{fig:prob}(a),
deep neural networks trained with the standard softmax Cross-Entropy (CE) loss tend to give overconfident predictions even when the test input differs from the training distribution~\cite{nguyen2015deep}. 
Motivate by this, 
our goal is to explicitly create a support region for unknown target samples. 
A native choice is the low-density regions with respect to the source-trained classifier. 

To empirically verify our intuitions, 
we use \emph{scikit-learn}~\cite{pedregosa2011scikit} to generate samples (3 known classes and 1 unknown class) and show the comparison in Fig.~\ref{fig:toy_example}. 
From the figure, we have the following observations.
(1) Simply training a ($|\mathcal{C}_s|$+1)-way classifier cannot improve the discrimination of unknown class. 
(2) Forcefully increasing the softmax probability in the unknown dimension creates an additional support region. 
However, due to the overconfidence issue regarding known classes, the response to the unknown (reflected by the size of the region) is still limited.    
(3) To increase the response to unknown class, 
we penalize the prediction confidence \emph{w.r.t.} known classes, \emph{i.e.,} making the known-class data closer to their decision boundaries.   
(4) Although the reshaped decision boundaries are able to accommodate unknown-class data, the boundaries between known and unknown classes are less discriminative as we do not have access to real unknown data, \emph{i.e.,} the unknown samples do not necessarily lie in the support of the created region since the above operations only encourage it far away from the support of known classes.
Thus, we dynamically adjust the learned boundaries using unlabeled test data.

Grounded on these insights, we propose a novel Activate and Reject~(\ar) approach.
Specifically, \ar\ encompasses two innovative components: 
1) Unknown-aware Gradient Diffusion~(UGD) to diffuse the gradient to the unknown's logit with smoothing regularization; 
2) Test-time Unknown Rejection~(TUR) to conduct post hoc modification to the learned classifier's final prediction.

\subsection{Unknown-aware Gradient Diffusion}
As discussed in Sec.~\ref{sec:motivation}, deep classifiers trained with the standard softmax CE loss are susceptible to the notorious overconfidence issue. This problem becomes more sophisticated in the context of DGCS, wherein the learned decision boundary is highly biased towards source known-class samples.
On the other hand, given only access to known-class data during the training phase, 
how to optimize the $|\mathcal{C}_s|$+1-way classifier is problematic (cf. Fig.~\ref{fig:toy_example}(a)).

With this premise, we propose the UGD to solve the above issues at training phase from two perspectives, 
\emph{i.e.,} unknown activation and output's smoothness.
The former activates the unknown probability,
while the latter mitigates the overconfidence issue. 
First of all, we need to train a ($|\mathcal{C}_s|$+1)-way classifier, where an additional dimension is introduced to discriminate unknown classes from known ones. 
Given $\mathbf{x}_s \in \mathcal{D}_s$ and a neural network $f(\mathbf{x};\theta)$ parameterized by $\theta$, we define the standard CE loss as:
\begin{equation}
	\mathcal{L}_{\text{CE}}(f(\mathbf{x_s}),y_s) 
	= -\log\frac{ \exp (f_{y_s}(\mathbf{x_s}))}{\sum_{k\in |\mathcal{C}_s|+1} \exp (f_k(\mathbf{x_s}))},
\end{equation}
where $f(\mathbf{x_s}) \in \mathbb{R}^{|\mathcal{C}_s|+1}$ denotes the network's logit and $f_{y_s}(\mathbf{x_s})$ is the $y_s$-th element of $f(\mathbf{x_s})$ corresponding to the ground-truth label $y_s$.

Based on the ($|\mathcal{C}_s|$+1)-way classifier, 
we aim to activate the unknown's logit in the absence of real unknown-class samples.
The key idea is to \emph{increase the value of unknown probability without affecting the ground-truth classification.} 
For notation shorthand, 
we use $\boldsymbol{f}_k$ to represent the logit of $k$-th class and $\boldsymbol{f}_u$ for the unknown's logit.
Since we have no supervision over the unknown, 
the value of $\boldsymbol{f}_u$ is negligible (cf. Fig.~\ref{fig:prob}(a)).
For a source sample $(\mathbf{x}_s,y_s) \in \mathcal{D}_s$,
we forcefully increase the unknown probability by minimizing the negative log-likelihood,
\begin{equation}\label{eq:UA}
	\mathcal{L}_{\text{UA}} = - \log \frac{ \exp (\boldsymbol{f}_u)}{\sum_{k\in |\mathcal{C}_s|+1, k \neq y_s} \exp (\boldsymbol{f}_k)},
\end{equation}
This objective ensures that the unknown probability can give a response to any input sample regardless of its class label (cf. gray region in Fig.~\ref{fig:toy_example}(b)).  
Since the learning process is always dominated by CE loss regarding the ground-truth category, Eq.~(\ref{eq:UA}) is tractable and will not hurt the known-class performance.  
However, the activated probability is relatively small (compared to the ground-truth category), 
which leads to an unsatisfactory accuracy for real unknown samples, 
especially for some hard samples (cf. Fig.~\ref{fig:prob}(b)). 

Next, we aim to enhance the response to unknown classes by increasing the smoothness of the network's output (cf. Fig.~\ref{fig:toy_example}(c)). 
Formally, we impose two constraints to the standard CE loss: a temperature scaling parameter $\tau$ ($\tau > 1$) and a penalty on the $L_2$ norm of the logits. 
Thus, the proposed smoothed CE (SCE) loss $\mathcal{L}_{\text{SCE}}$ is defined as:
\begin{equation}
	\mathcal{L}_{\text{SCE}} = -\log\frac{ \exp (f_{y_s}(\mathbf{x_s})/\tau)}{\sum_{i\in |\mathcal{C}_s|+1} \exp (f_i(\mathbf{x_s})/\tau)} + \lambda \lVert f(\mathbf{x_s}) \rVert_2,
\end{equation}
where $\lambda$ is set to 0.05 in all experiments. 

Finally, the UGD loss is formulated as:
\begin{align}
    \mathcal{L}_{\text{UGD}} = \mathcal{L}_{\text{UA}} + \mathcal{L}_{\text{SCE}}.
\end{align}
As shown in Fig.~\ref{fig:prob}(c), 
the proposed $\mathcal{L}_{\text{UGD}}$ not only reduces the overconfidence issue (smaller max-probability for known sample) but also significantly increases the unknown probability.

\subsection{Test-Time Unknown Rejection}
Although we have activated the network's logit about unknowns, 
there still exist two critical challenges that impede the safe and reliable deployment of our source-trained models on open-world data.
First, a conservative (smaller max-probability) and smoothing (larger entropy) output on source data may not guarantee the category correspondence across domains and therefore may lead to semantic misalignment.
Second, how to reject a sample as ``unknown'' lacks principled criterion considering that the unknown-class samples may distribute randomly in the embedding space.
In this regard, previous open-set-oriented methods~\cite{saito2020universal,zhu2022crossmatch} that typically rely on thresholding mechanisms (\emph{e.g.} entropy value~\cite{zhu2022crossmatch}) are heuristic and will be sensitive to the variations of domain disparity.

To solve the above issues, 
we introduce a simple and effective technique---TUR---to match unlabeled test data to the source-trained model in an online adaptation manner. 
Our key idea is to conduct \emph{post hoc} modification to the learned classifier's final predictions, so as to bring the decision boundaries of different classes closer to the well-behaved case.
TUR is \emph{training-free} (\emph{i.e.,} no backward passes) and does not impose any distributional assumptions.

Technically, 
we impose a cross-domain cycle-consistent constraint on the top of embedding space for identifying whether a test sample corresponds to any known classes or not. 
The cross-domain relationships are based on $K$-nearest neighbor (KNN)~\cite{johnson2019billion} to perform non-parametric density estimation, which is model-agnostic and easy to implement.
Specifically, we decompose the source-trained model into a feature extractor $g$ and a linear classifier $f$.
Assume that the embedding of training data is $\mathbb{Z}_s=\{\boldsymbol{z}_s^1,\boldsymbol{z}_s^2,...,\boldsymbol{z}_s^{n_s}\}$, where $\boldsymbol{z}_s^i$ is the $L_2$-normalized penultimate feature $\boldsymbol{z}_s^i=g(\mathbf{x}_s)/\lVert g(\mathbf{x}_s) \rVert_2$. 
Here, we do not require access to the original training samples since the embedding will be extracted in advance, and no need to update.  
Then, we define two sets of known-class prototypes on the top of penultimate layer, \emph{i.e.,} $\{\mu_s^k\}_{k=1}^{|\mathcal{C}_s|}$ and $\{\mu_t^k\}_{k=1}^{|\mathcal{C}_s|}$,
where $\mu_s^k$ is computed from $\mathbb{Z}_s$ (mean feature per class) and will be fixed at test time.
$\mu_t^k$ is empty at the beginning.

For an test input $\mathbf{x}_t^j$ with its normalized feature vector $\boldsymbol{z}_t^j$,
we compute its KNN in $\mathbb{Z}_s$, denoted by $\mathcal{N}_s(\boldsymbol{z}_t^j)$. 
The feature centroid of $\mathcal{N}_s(\boldsymbol{z}_t^j)$ is denoted by $\boldsymbol{\bar{z}}_s^j$.    
Next, we find the corresponding source class as,
\begin{align}
\vspace{-0.2cm}
	k' = \argmax_{k' \in \{0,1, \cdots,|\mathcal{C}_s|\}}{sim(\boldsymbol{\bar{z}}_s^j, \mu_s^{k'})}
\vspace{-0.2cm}
\end{align}
Here, we measure the cosine similarity between features as: 	$sim(\boldsymbol{\bar{z}}_s^j, \mu_s^{k'}) = \frac{(\boldsymbol{\bar{z}}_s^j)^T  \mu_s^{k'}}{\lVert \boldsymbol{\bar{z}}_s^j \rVert_2  \lVert \mu_s^{k'} \rVert_2}$.
In the same way, we search the target class $k''$ based on the similarity between $\boldsymbol{\bar{z}}_s^j$ and $\mu_t^{k''}$. 
If $k'$ and $k''$ belongs to the same category, the sample $\boldsymbol{x}_t^j$ will be predicted as class $k''$ and we further update $\mu_t^{k''}$ in the following manner,
\begin{equation}\label{eq:target}
    \mu_{t(I)}^{k''} = {\phi\,\boldsymbol{z}_t^j+(1-\phi)\,\mu_{t(I)}^{k''}},
\end{equation}
where $\mu_{t(I)}^{k''}$ denote the $k''$-th target prototype until time $I$ and $\phi \in (0,1)$ is a preset scalar and fixed to 0.3 in practice. 

%
%
If $k'$ and $k''$ belong to different categories, 
the prediction will be given by using a follow-up strategy. 
Specifically, 
a memory bank $\mathbb{M}_I=\{\mathbb{M}^1_I, \cdots, \mathbb{M}^{|\mathcal{C}_s|+1}_I\}$ is a set of target sample embedding until time $I$, which is initialized by the weight of linear classifier $f$. 
At time $I$, $\mathbb{M}_I$ is updated as:
\begin{align}
	\mathbb{M}^k_I = 
	\begin{cases}
		\mathbb{M}^k_{I-1} \cup \boldsymbol{z}_t^j  & \text{if}\;k' \neq k''\,\text{and}\,f(\boldsymbol{z}_t^j) = k, \\
		\mathbb{M}^k_{I-1} & \text{otherwise},
	\end{cases} 
\end{align}
Similarly, we can build a new set of target class prototypes $\{\psi_t^k\}_{k=1}^{|\mathcal{C}_s|+1}$ based on samples from $\mathbb{M}_I$. Note that $\psi_t^k$ will be constantly updated during test time. Then, we predict the class label (($|\mathcal{C}_s|$+1)-way) of $\mathbf{x}_t^j$ as follows, 
\begin{align}
\vspace{-0.2cm}
\hat{k} = \argmax_{\hat{k} \in \{0,1, \cdots,|\mathcal{C}_s|+1\}}{sim(\boldsymbol{\bar{z}}_s^j, \psi_t^{\hat{k}})}.
\vspace{-0.2cm}
\end{align}
The decision boundaries between known and unknown classes are refined without backpropagation (cf. Fig.~\ref{fig:toy_example}(d)). 

\begin{center}
\begin{table*}[t]
        \centering
        \setlength\tabcolsep{5pt} 
	\caption{Accuracy (\%) on four classification benchmarks (ResNet-18).} 
	\label{tab:cls}
	\scalebox{0.78}{
	\begin{tabular}{ccccacccacccacccaccca} 
        \toprule
    \multirow{2}{*}{Regime} & \multirow{2}{*}{Method} & \multicolumn{3}{c}{\textbf{PACS}} && \multicolumn{3}{c}{\textbf{Office-Home}} && \multicolumn{3}{c}{\textbf{Office-31}} && \multicolumn{3}{c}{\textbf{Digits}} && \multicolumn{3}{c}{\textbf{Average}}  \\ 
			\cmidrule{3-5} \cmidrule{7-9} \cmidrule{11-13} \cmidrule{15-17} \cmidrule{19-21}
		&	& $acc_k$  & $acc_u$  & $hs$  && $acc_k$  & $acc_u$  & $hs$              && $acc_k$  & $acc_u$  & $hs$             &&$acc_k$  & $acc_u$  & $hs$                 && $acc_k$  & $acc_u$  & $hs$              \\ 
			\midrule
            \multirow{2}{*}{\makecell{OSDA \\ (\small upper bound)}} & OSBP~\cite{saito2018open} & 40.6 & 49.5 & 44.6 && 47.1 & 66.9 & 55.3 && 75.8 & 84.3 & 77.7 && 35.6 & 70.6 & 40.5 && 49.8 & 67.8 & 54.5\\
            & ROS~\cite{bucci2020effectiveness}  & 35.6 & 66.4 & 46.4 && 50.8 & 77.5 & 60.8 && 71.7 & 80.0 & 75.6 && 20.1 & 48.6 & 34.9 && 47.7  &  68.1 & 54.4\\
            \midrule
            \multirow{3}{*}{OD} & 
            MSP~\cite{hendrycks2017baseline} & 38.9 & 62.5 & 46.4 && 52.7 & 75.6 & 62.0 && 49.7 & 89.2 & 63.8 && 17.2 & 87.1 & 28.8 && 39.6 & 78.6 & 50.3\\
            & LogitNorm~\cite{wei2022mitigating} & 35.1 & 47.6 & 38.3 && 56.3 & 56.5 & 56.1 && 41.0 & 71.2 & 52.1 && 26.8 & 51.2 & 35.2 &&  39.8 & 56.6 & 45.4 \\
            & DICE~\cite{sun2022dice} & 44.0 & 53.4 & 49.2 && 61.5 & 58.8 & 59.9 && 72.8 & 61.1 & 66.4 && 35.0 & 47.6 & 40.3 && 53.3 & 55.2 & 54.0 \\
            \midrule
            \multirow{2}{*}{SFDA} & 
            SHOT~\cite{liang2020we} & 51.2 & 34.9 & 40.8 && 52.5 & 32.4 & 44.3 && 84.8 & 60.2 & 70.4 && 27.4 & 20.3 & 23.3 && 54.0 & 37.0  & 44.7\\
            & AaD~\cite{yang2022attracting} & 45.1 & 40.0 & 42.0 && 59.4 & 58.7 & 58.9 && 70.1 & 85.3 & 76.9 && 25.6 & 26.9 & 26.2 && 50.1 & 52.7 & 51.0\\
            \cmidrule{2-21}
            \multirow{3}{*}{TTA} 
            & TTT~\cite{sun2020test} & 36.9 & 44.6 & 38.9 && 52.0 & 45.9 & 47.2 && 35.4 & 79.6 & 49.0 && 44.1 & 45.1 & 44.6 && 42.1 & 53.8 & 44.9\\
            & Tent~\cite{wang2021tent} & 25.2 & 43.1 & 31.7 && 33.6 & 45.9 & 38.7 && 56.0 & 85.1 & 67.5 && 27.2 & 41.1 & 32.7 && 35.5 & 53.8 & 42.7\\
            & MEMO~\cite{zhang2022memo} & 37.9 & 52.3 & 44.5 && 49.0 & 55.6 & 52.1 && 59.8 & 72.7 & 65.6 && 21.7 & 56.1 & 31.3 &&  42.1 & 59.2 & 48.4\\
            \midrule
            \multirow{7}{*}{OSDG} & ERM~\cite{vapnik1999overview} & 52.3 & 27.0 & 36.1 && 66.9 & 23.7 & 34.3 && 85.1 & 27.0 & 40.7 && 56.4 & 13.0 & 18.0 && 65.2 & 22.7 & 32.3\\
	    & ADA~\cite{volpi2018generalizing} & 54.2 & 30.9 & 36.4 && 67.9 & 25.4 & 36.2 && 85.6 & 25.2 & 38.7 && 57.2 & 15.1 & 20.1 && 66.2 & 24.2 & 32.9\\
		& ADA+CM~\cite{zhu2022crossmatch} & 56.4 & 45.6 & 43.0 && 65.0 & 40.4 & 48.5 && 83.0 & 34.5 & 48.5 && 49.2 & 52.1 & 39.9 && 63.4 & 43.2 & 45.0\\
		& MEADA~\cite{zhao2020maximum} & 54.1 & 31.4 & 36.2 && 67.6 & 25.7 & 36.4 && 85.8 & 25.1 & 38.6 && 57.6 & 29.8 & 30.4 && 66.3 & 28.0 & 35.4\\
		& MEADA+CM~\cite{zhu2022crossmatch} & 54.3 & 46.6 & 42.7 && 64.9 & 40.5 & 49.6 && 82.8 & 41.1 & 54.7 && 52.3 & 46.1 & 38.7 && 63.6 & 43.6 & 46.4\\ 
		&	One Ring-S~\cite{yang2022one}  & 43.7 & 49.4 & 41.5 && 56.9 & 69.0 & 62.3 && 67.3 & 77.0 & 71.3 && 33.2 & 51.3 & 40.3 && 50.3 & 61.7 & 53.9 \\
        & \ar\ w/o TUR & 47.0 & 51.3 & 48.1 && 58.8 & 69.8 & 63.7 && 70.7 & 65.9 & 68.2  &&  29.7 &  65.7 & 40.9 && 51.6 & 63.2 & 55.2\\
        \cmidrule{2-21}
	DGCS & \ar~(full) &  43.7 & 65.9 & \textbf{52.3} && 64.3 & 65.3 & \textbf{64.8} && 82.1 & 75.2 & \textbf{78.5}  && 34.3 & 63.8 & \textbf{44.6} && 56.1 & 67.6 & \textbf{60.1} \\
	\bottomrule 
        \end{tabular}}
\end{table*}
\end{center}

\begin{center}
	\begin{table*}[t]	
		\caption{Performance of \ar\ on object detection benchmarks.} 
        \label{tab:detection}
		\centering 
		\scalebox{0.8}{
			\begin{tabular}{l|ccccc|ccccc|ccccc} 
				\toprule
				\multirow{2}{*}{Method} & \multicolumn{5}{c|}{\textit{Pascal VOC}$\rightarrow$\textit{Clipart}} & \multicolumn{5}{c|}{\textit{Pascal VOC}$\rightarrow$\textit{Watercolor}} & \multicolumn{5}{c}{\textit{Pascal VOC}$\rightarrow$\textit{Comic}}  \\ 
			\cmidrule{2-16}
				& WI$_\downarrow$  & AOSE$_\downarrow$  & mAP$_\mathcal{K \uparrow}$ & AP$_\mathcal{U \uparrow}$& $hs_\uparrow$ & WI$_\downarrow$  & AOSE$_\downarrow$  & mAP$_\mathcal{K \uparrow}$ & AP$_\mathcal{U \uparrow}$& $hs_\uparrow$ & WI$_\downarrow$  & AOSE$_\downarrow$  & mAP$_\mathcal{K \uparrow}$ & AP$_\mathcal{U \uparrow}$& $hs_\uparrow$ \\ 
			\midrule
                ORE~\cite{joseph2021towards} & 17.3 & 876 & \textbf{37.7} & 3.0 & 5.6 & 28.4 & 3216 & 19.8 & 13.5 & 16.1 & 23.1 & 2242 & {7.3} & 3.0 & 4.3\\
                OpenDet~\cite{han2022expanding} & {14.2} & \textbf{300} & 32.7 & {6.7} & {11.1} & \textbf{14.9} & 1944 & 19.2 & \textbf{19.3} & {19.2} & 15.2 & {744} & \textbf{7.5} & 3.1 & 4.4\\
                \midrule
                \ar\ (full) & \textbf{11.7} & {317} & {35.8} & \textbf{10.2} & \textbf{15.9} & {19.7} & \textbf{944} & 20.8 & 15.2 & 17.6 & \textbf{13.2} & \textbf{596} & 7.2 & \textbf{9.1} & \textbf{8.0} \\
                 w/o $\mathcal{L}_{\text{UA}}$ & 16.3 & 1363 & 35.4 & 6.0 & 10.3 & 29.4 & 3924 & 18.6 & 14.1 & 16.0 & 25.0 & 2826 & 6.4 & 2.2 & 3.3\\
				w/o $\mathcal{L}_{\text{SCE}}$ & 14.6 & 426 & 34.7 & 3.2 & 5.9 & 24.8 & {1104} & \textbf{21.4} & {19.1} & \textbf{20.2} & 24.7 & 1372 & 6.3 & 3.5 & 4.5\\
				w/o TUR & 14.9 & 444 & 34.5 & 4.9 & 8.6 & 23.3 & 1398 & {21.1} & 17.6 & {19.2} & {15.0} & 784 & {7.3} & {4.6} & {5.6}\\
				\bottomrule
		\end{tabular}}
	\end{table*}
\end{center}

\begin{center}
	\begin{table}[t]
		\centering
		\caption{Performance of \ar\ on semantic segmentation benchmark, \emph{i.e.,} from \textbf{GTA5} (synthetic) to \textbf{Cityscapes} (real).} 
	\small
	\label{tab:segmentation}
		\begin{tabular}{lcccca} 
			\toprule
			Method && mAcc & mIOU & $acc_u$ & $hs$ \\ 
			\midrule
			   ERM~\cite{vapnik1999overview}  && 64.9 & 48.2 & 27.6 & 39.4\\
                One Ring-S~\cite{yang2022one} && 55.7 & 41.0 & 72.5 & 61.9 \\
                \midrule
                \ar\ (full) && 57.1 & 43.3 & 73.2 & \textbf{63.1} \\
			w/o UGD  && 64.2 & 46.6 & 41.6 & 50.2\\
			w/o TUR  && 54.7 & 42.6 & 78.5 & 62.6 \\
			\bottomrule
		\end{tabular}
	\end{table}
\end{center}

\vspace{-28mm}
\section{Experiments}


\subsection{Generalization in Image Classification}
\noindent
\textbf{Dataset.} We evaluate our \ar\ on four standard DG benchmarks.
\textbf{PACS}~\cite{li2017deeper}, which has dramatic differences in terms of image styles, contains 9,991 images of seven object classes from four domains, \emph{i.e.,} \emph{Photo}, \emph{Art Painting}, \emph{Cartoon}, and \emph{Sketch}. 
4 classes (dog, elephant, giraffe, and guitar) are adopted as $\mathcal{C}_s$ and the remaining 3 classes are used as $\mathcal{C}_u^t$.
\textbf{Office-Home}~\cite{venkateswara2017deep}, which is collected from office and home environments, 
has 15,500 images of 65 classes from four domains, \emph{i.e.,} \emph{Artistic}, \emph{Clipart}, \emph{Product}, and \emph{Real World}. The domain shifts stem from the variations of viewpoint and image style. 
In alphabetic order, the first 15 classes are selected as $\mathcal{C}_s$ and the remaining 50 classes are used as $\mathcal{C}_u^t$.
\textbf{Office-31}~\cite{saenko2010adapting} has 31 classes collected from three domains: \emph{Amazon}, \emph{DSLR}, and \emph{Webcam}. 
The 10 classes shared by Office-31 and Caltech-256~\cite{gong2012geodesic} are adopted as $\mathcal{C}_s$. 
In alphabetical order, the last 11 classes along with $\mathcal{C}_s$ form $\mathcal{C}_u^t$.  
\textbf{Digits}, which differs in the background, style, and color, contains four handwritten digit domains including \emph{MNIST}~\cite{lecun1998gradient}, \emph{MNIST-M}~\cite{ganin2015unsupervised}, \emph{SVHN}~\cite{netzer2011reading}, \emph{USPS}~\cite{hull1994database}, and \emph{SYN}~\cite{ganin2015unsupervised}.
\emph{MNIST} is used as the source domain and the other datasets are viewed as target domains.
$\mathcal{C}_s$ includes numbers from 0 to 4.

\noindent
\textbf{Evaluation Protocols.}
Following~\cite{bucci2020effectiveness,zhu2022crossmatch,yang2022one}, we adopt H-score ($hs$)~\cite{fu2020learning} as the main evaluation metric. 
$hs$ harmonizes the importance of known and unknown classes by requiring that known and unknown class accuracy should be both high and balanced.
The known class accuracy ($acc_k$) and unknown class accuracy ($acc_u$) are also provided. 

\noindent
\textbf{Implementation Details.} 
We conduct experiments based on Dassl~\cite{Dassl}, 
including data preparation, model training, and model selection.
For PACS, Office-Home, and Office-31,
we use ResNet-18~\cite{he2016deep} pre-trained on the ImageNet as the backbone network.
We use the ConvNet~\cite{lecun1989backpropagation} with architecture \emph{conv-pool-conv-pool-fc-fc-softmax} for Digits. The networks are trained using SGD with momentum of 0.9 for 100 epochs.
The batch size is set to 16. 

\noindent
\textbf{Baselines.}
Given the contact points with other problem settings,
we compare \ar\ with five types of state-of-the-art methods.
(1) \textbf{OSDG}~\cite{zhu2022crossmatch,yang2022one} is the most related baseline. \textit{When TUR is removed, the proposed \ar\ becomes a standard OSDG method.} 
(2) \textbf{OSDA}~\cite{saito2018open,bucci2020effectiveness} jointly utilizes source and target data for training and thus can be viewed as an upper bound of our problem. 
(3) \textbf{OD}~\cite{wei2022mitigating,sun2022dice} usually identifies unknown-class samples via scoring functions. 
(4) \textbf{SFDA}~\cite{liang2020we,yang2022attracting} and \textbf{TTA}~\cite{liang2020we,yang2022attracting} cannot deal with unknown-class samples directly. Therefore, we follow~\cite{zhu2022crossmatch} that uses the entropy of softmax output as the normality score. 


\noindent
\textbf{Results.}
The classification results on PACS and Office-Home, Office-31, and Digits benchmarks are reported in Tab.~\ref{tab:cls}.
\ar\ substantially and consistently outperforms baseline methods on different benchmark datasets. 
For example, \ar\ improves $hs$ by 9.3\% (PACS), 2.5\% (Office-Home), 7.2\% (Office-31), and 4.3\% (Digits) compared to the previous best OSDG baselines. 
In particular, only using UGD could also substantially exceed state-of-the-art methods, \emph{e.g.} DICE~\cite{sun2022dice} and One Ring-S~\cite{yang2022one}. 
The results also reveal several interesting observations.
(1) The performance of~\cite{yang2022one} is unstable across different benchmarks. For example, they outperform CM~\cite{zhu2022crossmatch} by +12.7\% and +16.6\% on Office-Home and Office-31 but show inferior performance (-1.5\%) on PACS.
By contrast, our method achieves more consistent improvements, indicating the efficacy and scalability of \ar.
(2) \ar\ achieves even better performance than OSDA methods (upper bound) under much more challenging settings. 
(3) LogitNorm~\cite{wei2022mitigating} and Tent~\cite{wang2021tent} achieve inferior performance due to the imbalance between $acc_k$ and $acc_u$, showing the non-triviality of performing both unknown-aware training and test-time modification.  
 

\subsection{Generalization in Other Vision Tasks}
\noindent
\textbf{Setup.}
\textbf{(1) Object Detection.} We introduce four datasets to form three tasks, \emph{i.e.,} Pascal VOC~\cite{everingham2010pascal}, Clipart, Watercolor, and Comic~\cite{inoue2018cross} datasets. 
They share 6 classes, where \emph{person} is selected as $\mathcal{C}_u^t$ and the remaining 5 classes are viewed as $\mathcal{C}_s$. 
The Pascal VOC2007-trainval and VOC2012-trainval datasets are combined to form the source domain, and Clipart1k, Watercolor, and Comic as used as the target domains respectively. 
For evaluation, we introduce four metrics:
Wilderness Impact (WI)~\cite{dhamija2020overlooked}, Absolute Open-Set Error (AOSE)~\cite{miller2018dropout}, mean average precision of known classes (mAP$_\mathcal{K}$) and average precision of unknown class (AP$_\mathcal{U}$). 
\textbf{(2) Semantic Segmentation.}
GTA5~\cite{richter2016playing} and Cityscapes~\cite{cordts2016cityscapes} are used as the source and target domains respectively.
GTA5 is a synthetic dataset generated from Grand Theft Auto 5 game engine, while Cityscapes is collected from the street scenarios of different cities. 
They share 19 classes in all.
According to the number of pixels per class, we use 10 classes as $\mathcal{C}_s$ and the remaining 9 classes as $\mathcal{C}_t^u$. 
We report the mean accuracy of all classes (mAcc), mean Intersection over Union (mIOU), $acc_u$ and $hs$. 

\noindent
\textbf{Implementation Details.} 
\textbf{(1) Object Detection.} We utilize Faster R-CNN~\cite{ren2015faster} as the detection model and ResNet-50 with FPN~\cite{lin2017feature} as the backbone network. To avoid the mutual influence between classification and regression heads, the original shared FC layer is replaced by two parallel FC layers. The networks are trained for 40 epochs.
\textbf{(2) Semantic Segmentation.} We adopt DeepLab-v2~\cite{chen2017deeplab} segmentation network with ResNet-101 backbone. We use SGD optimizer with an initial learning rate of $5 \times 10^{-4}$, momentum of 0.9, and weight decay of $10^{-4}$.

\begin{figure*}[!h]
	\centering
	\includegraphics[width=1\textwidth]{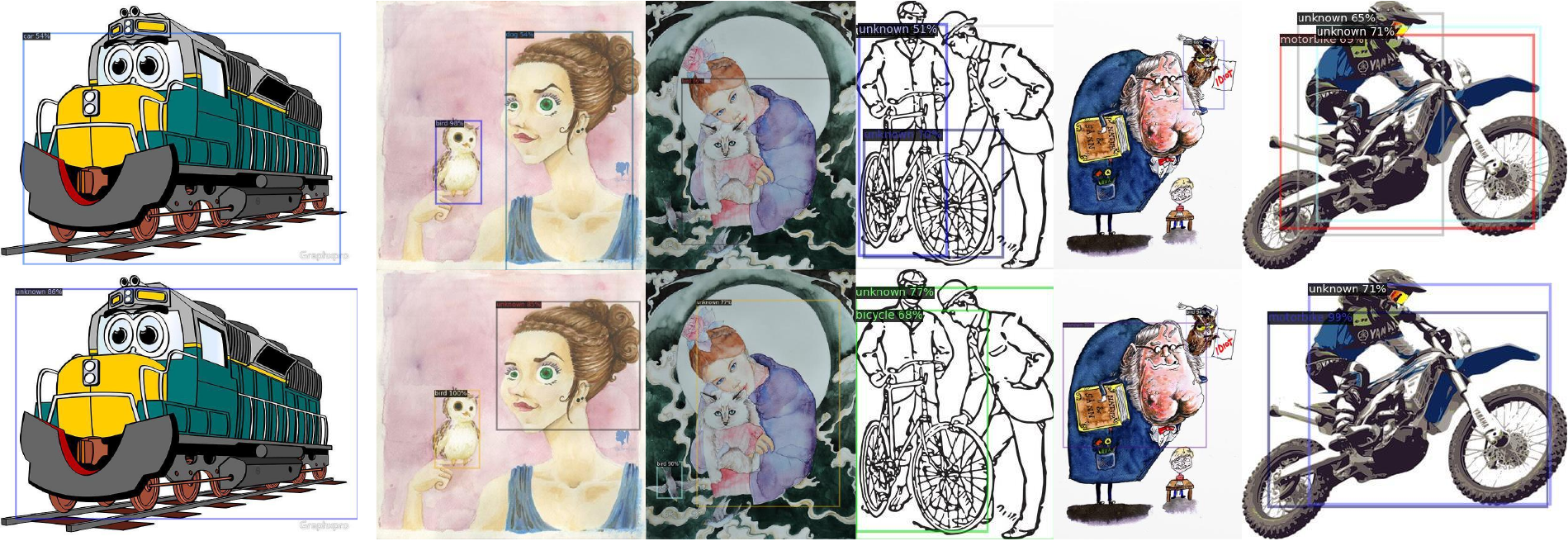}
	\caption{Qualitative comparisons between OpenDet (top) and \ar\ (bottom).}
	\label{fig:det}
\end{figure*}
\begin{figure*}[!h]
	\centering
	\includegraphics[width=1\textwidth]{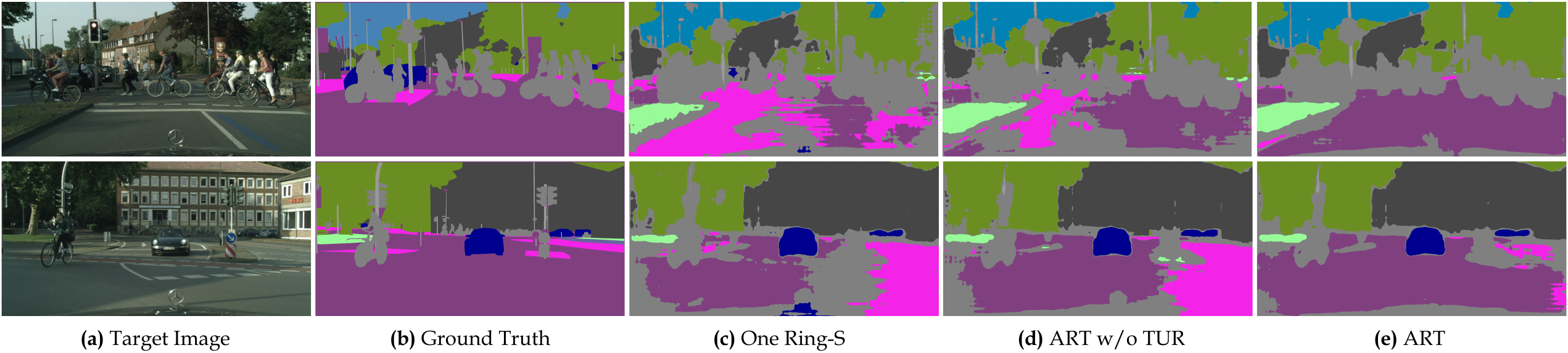}
	\caption{Visualization of segmentation results for the task GTA5 $\rightarrow$ Cityscapes. Gray regions indicate the unknown-class pixels.}
	\label{fig:seg}
\end{figure*}

\noindent
\textbf{Results.}
Tab.~\ref{tab:detection} shows the detection results compared to ORE~\cite{joseph2021towards}, OpenDet~\cite{han2022expanding}, and several variants of \ar. 
With respect to mAP$_\mathcal{K}$ and AP$_\mathcal{U}$, \ar\ outperforms the previous best method by 1.5\% and 1.8\% on average, revealing that \ar\ strikes a better balance between identifications of known- and unknown-class objects. 
Fig.~\ref{fig:det} provides the qualitative comparisons, where \ar\ could precisely identify unknown samples and exhibits better bounding box regression results.
For semantic segmentation,
Tab.~\ref{tab:segmentation} reveal that even in the dense prediction task, \ar\ is capable of significantly improving the generalization ability of deep models.
The qualitative results are shown in Fig.~\ref{fig:seg}, where the predictions given by \ar\ are smoother and contain much fewer spurious areas than One Ring-S~\cite{yang2022one} and \ar\ w/o TUR, especially on the unknown classes (\emph{rider} and \emph{bike}).


\begin{figure*}[!h]
	\centering
	\small
	\setlength\tabcolsep{1mm}
	\renewcommand\arraystretch{0.1}
	\begin{tabular}{ccccc}
		\includegraphics[width=0.18\linewidth]{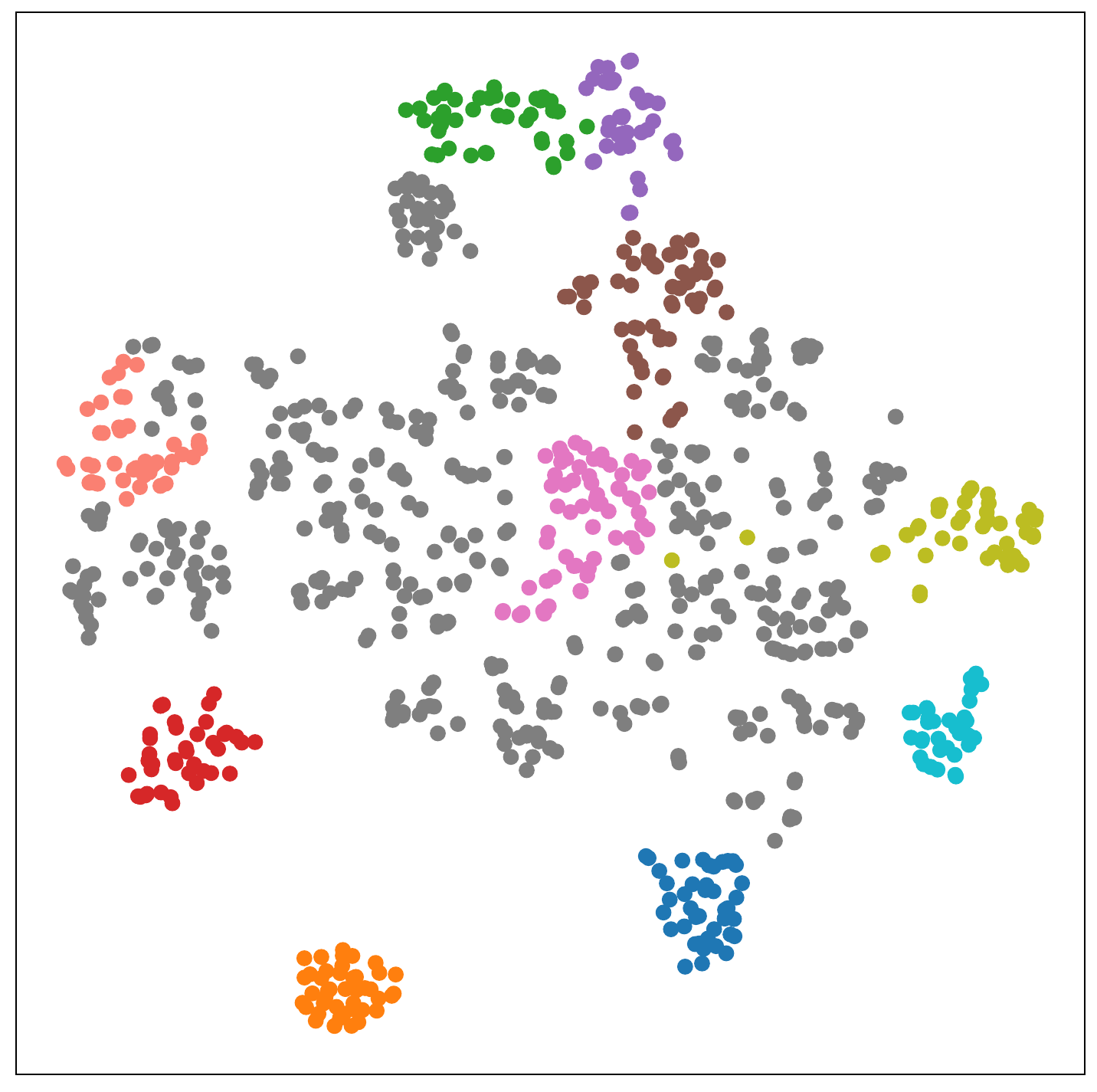} &
		\includegraphics[width=0.18\linewidth]{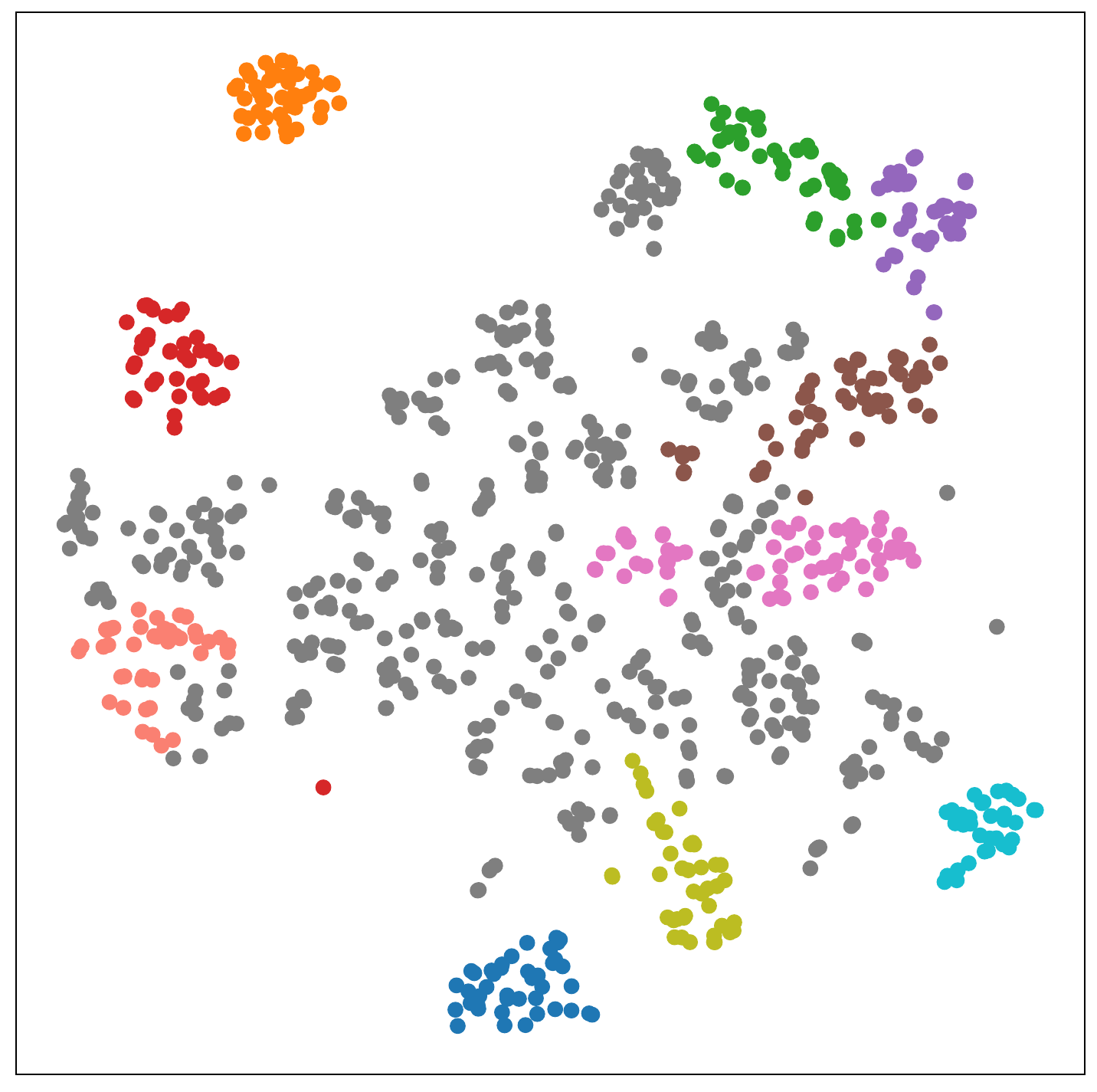} & 
		\includegraphics[width=0.18\linewidth]{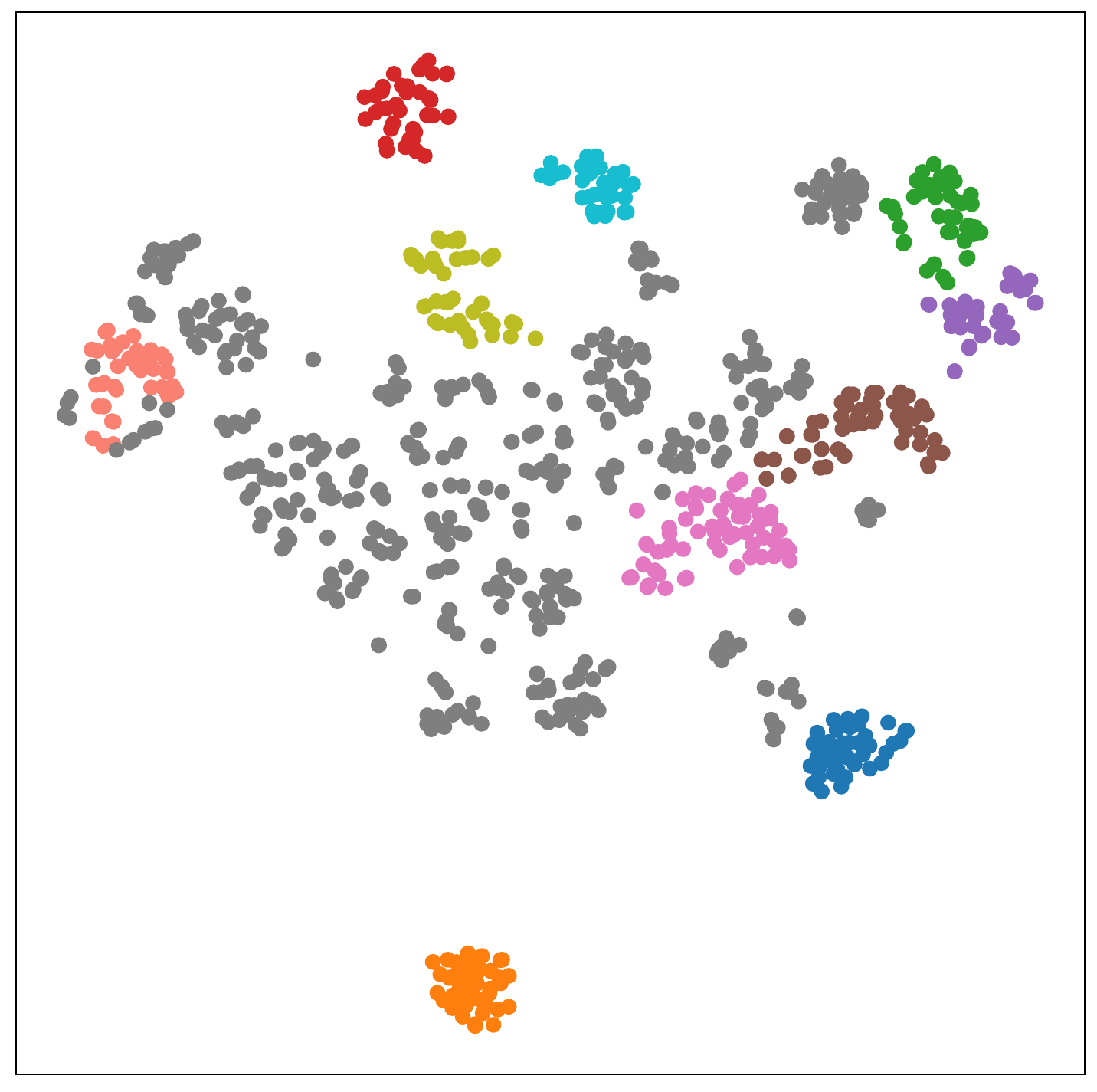} &
		\includegraphics[width=0.18\linewidth]{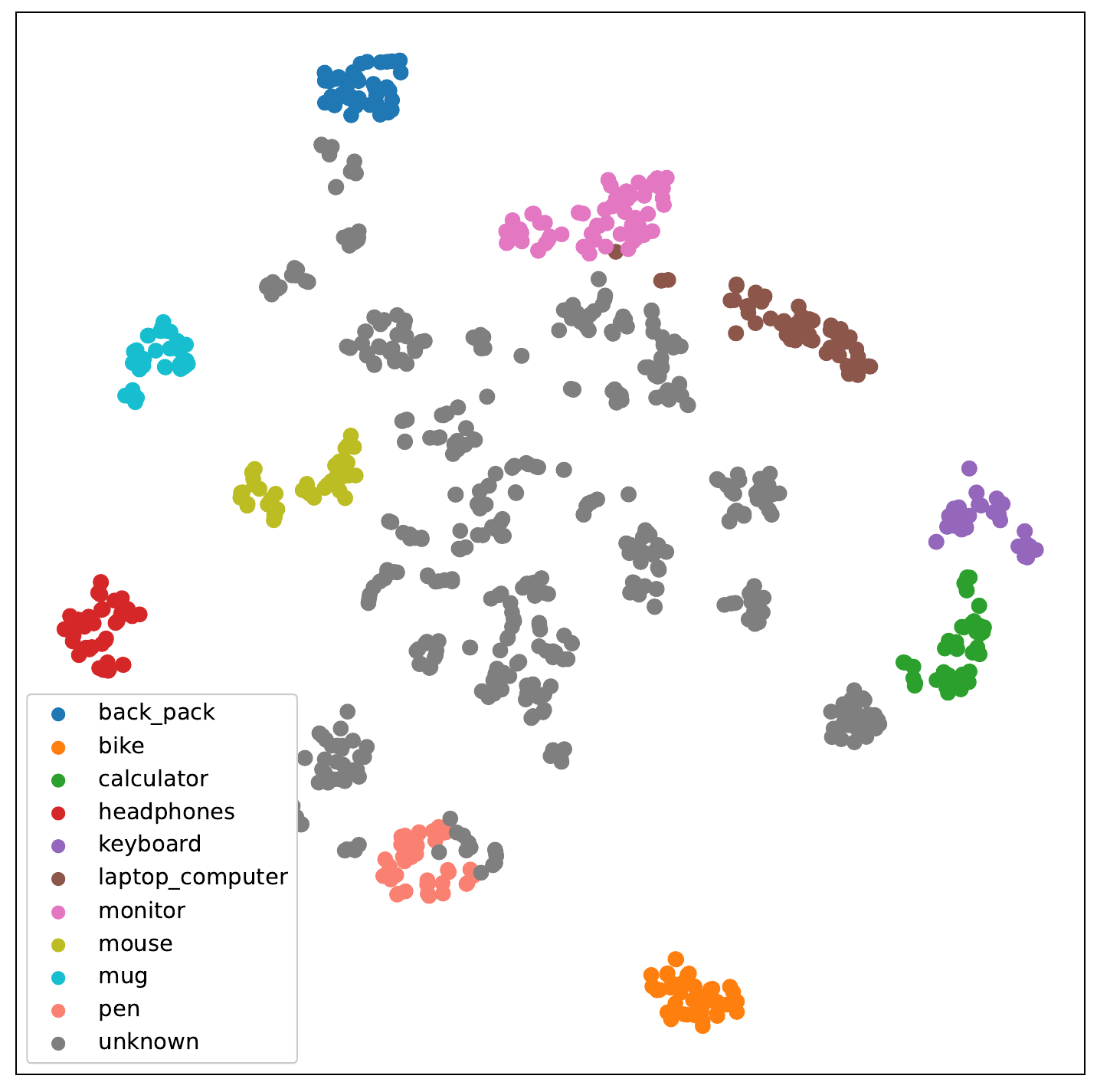} &
		\includegraphics[width=0.25\linewidth]{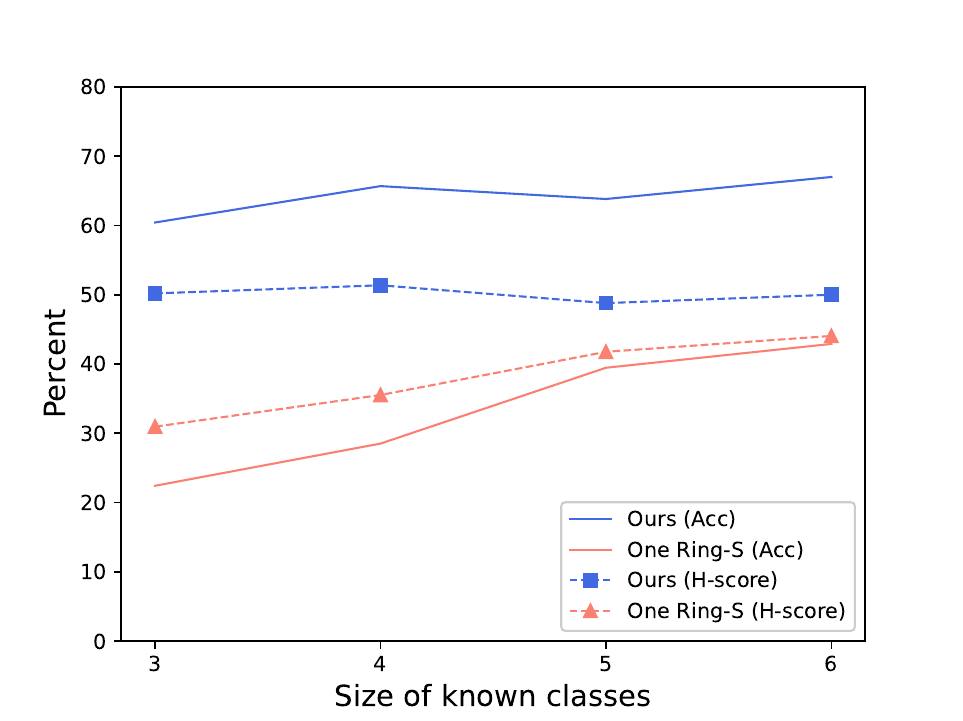} \\
		(a) ERM  & (b) One Ring-S & (c) \ar\ w/o $\mathcal{L}_{\text{SCE}}$ & (d) \ar\ & (e) One Ring-S \emph{vs.} \ar\   \\
	\end{tabular}
        \vspace{1mm}
	\caption{(a)-(d) t-SNE visualization~\cite{van2008visualizing} of the penultimate layer's feature on Office-31. (e) Varying the size of known classes on PACS.}
	\label{fig:tnse}
\end{figure*} 

\begin{figure}[t]
	\centering
	\includegraphics[width=0.45\textwidth]{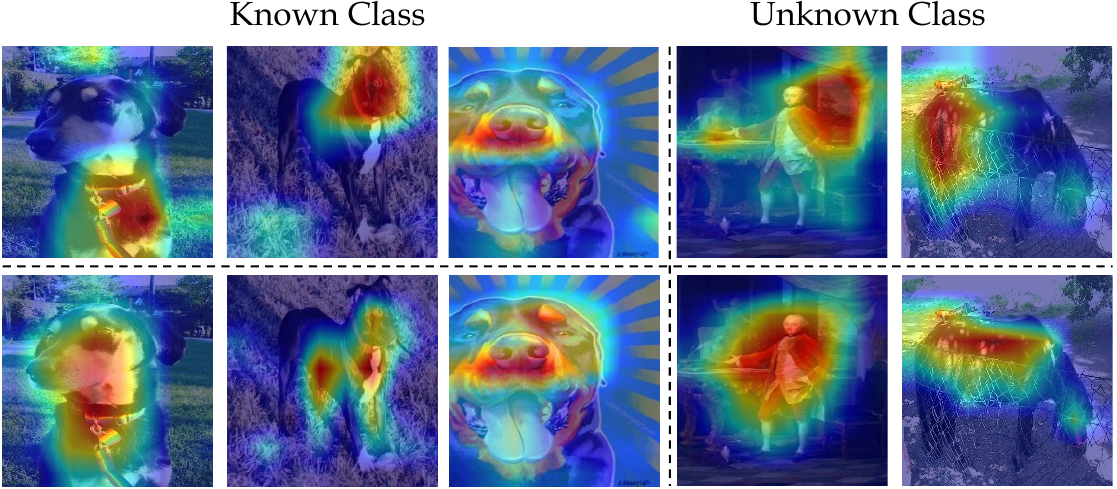}
	\caption{\textbf{Top:} w/o $\mathcal{L}_{\text{SCE}}$ \emph{vs.} \textbf{Bottom:} w/ $\mathcal{L}_{\text{SCE}}$}
	\label{fig:CAM}
	\vspace{-5mm}
\end{figure}

\begin{center}
	\begin{table}[t]
		\caption{Ablation of \ar\ on four benchmarks. $hs$ (\%) is reported. 
  \textcolor{red}{-} and \textcolor{citecolor}{+} denote the removal or addition of a module respectively.}
		\centering
            \small
		\setlength\tabcolsep{5pt}
		\label{tab:ablation}
            \scalebox{0.9}{
		\begin{tabular}{lccccc}
			\toprule
			Method & PACS & Office-Home & Office-31 & Digits & Avg. \\
			\midrule
                \rowcolor{ggray} \ar\ & 52.3 & 64.8 & 78.5 & 44.6 & 60.1 \\
                \midrule
                \textcolor{red}{-} $\mathcal{L}_{\text{UA}}$ & 39.9 & 62.5 & 69.0 & 20.0 & 47.9 \\
                \textcolor{red}{-} $\mathcal{L}_{\text{SCE}}$ & 43.4 & 60.0 & 71.5 & 41.3 & 54.1 \\
                \textcolor{red}{-} UGD & 45.5 & 57.0 & 66.6 & 32.0 & 50.3 \\
                \textcolor{red}{-} TUR \& $\mathcal{L}_{\text{UA}}$ & 44.4 & 61.4 & 65.8 & 7.9 & 44.9 \\
                \textcolor{red}{-} TUR \& $\mathcal{L}_{\text{SCE}}$ & 41.0 & 58.9 & 63.0 & 40.3 & 50.8 \\
		    \midrule
                \midrule
                \rowcolor{ggray} UGD & 48.1 & 63.7 & 68.2 & 40.9 & 55.2 \\
                \midrule
                \textcolor{citecolor}{+} TTT~\cite{sun2020test} & 48.5 & 60.8 & 72.8 & 41.3 & 55.9 \\
                \textcolor{citecolor}{+} Tent~\cite{wang2021tent} & 37.8 & 45.3 & 64.9 & 33.2 & 45.3 \\
                \textcolor{citecolor}{+}
                T3A~\cite{iwasawa2021test} & 49.2 & 62.7 & 72.0 & 41.7 & 56.4 \\
                \textcolor{citecolor}{+} MEMO~\cite{zhang2022memo} & 49.9 & 61.4 & 75.4 & 41.0 & 56.9 \\
                \textcolor{citecolor}{+} 
                SHOT~\cite{liang2020we} & 46.6 & 50.3 & 71.5 & 33.5 & 50.5 \\
                \textcolor{citecolor}{+} 
                AaD~\cite{yang2022attracting} & 50.2 & 62.5 & 74.7 & 41.8 & 57.3 \\
            \bottomrule
		\end{tabular}}
	\end{table}
\end{center}

\begin{center}
	\begin{table}[t]
		\caption{The influence of the order of test data. $hs$ is reported.}
		\vspace{1mm}
		\centering
            \small
		\setlength\tabcolsep{5pt}
		\label{tab:order}
            \scalebox{0.9}{
		\begin{tabular}{lccccc}
			\toprule
			ID & PACS & Office-Home & Office-31 & Digits & Avg. \\
			\midrule
                1 & 52.5 & 64.8 & 78.2 & 44.4 & 60.0 \\
                2 & 52.3 & 64.6 & 78.9 & 45.0 & 60.2 \\
                3 & 52.7 & 64.9 & 78.8 & 44.8 & 60.3 \\
                4 & 52.2 & 64.7 & 78.4 & 44.7 & 60.0 \\
            \bottomrule
		\end{tabular}}
	\end{table}
\end{center}

\vspace{-1.8cm}

\subsection{Discussion}

\noindent
\textbf{Ablation study.}
(1) In Tab.~\ref{tab:ablation}, we evaluate the contribution of the different components of \ar. 
It is evident that each of these components is reasonably designed, as the removal of any one of them leads to a commensurate reduction in accuracy. 
Note that when $\mathcal{L}_{\text{UA}}$ is removed, we will make the prediction by following the thresholding mechanism in~\cite{zhu2022crossmatch}.
To isolate the contribution of TUR, 
We additionally combine UGD with different TTA and SFDA methods. 
Notably, Tent and SHOT achieve inferior performance, and TTT and MEMO bring marginal improvements compared to the proposed TUR. 
(2) In fig.~\ref{fig:CAM}, we use Grad-CAM~\cite{zhou2016learning} to visualize the results trained w/ and w/o $\mathcal{L}_{\text{SCE}}$ on both target known- and unknown-class samples. We can observe that $\mathcal{L}_{\text{SCE}}$ makes the network focus on the entire object rather than a small or inaccurate local region, revealing the importance of mitigating the overconfidence issue in DGCS tasks. 









\noindent
\textbf{The influence of known classes.}
With fixed $|\mathcal{C}_s \cup \mathcal{C}_t|$, we investigate the influence of the number of known classes. As shown in Fig.~\ref{fig:tnse}, \ar\ consistently outperforms the previous best method in terms of $hs$ especially when the size is small, indicating that \ar\ can improve the generalization ability even with very limited known knowledge.  

\noindent
\textbf{The influence of test order.} As TUR is performed online, we study the influence of the order of test data. The results in Tab.~\ref{tab:order} reveal that TUR is insensitive to the variations of data order, showing its robustness to the open world.  

\noindent
\textbf{Feature visualization.}
We use t-SNE~\cite{van2008visualizing} to visualize the feature learned by ERM, One Ring-S, \ar\ w/o $\mathcal{L}_\text{SCE}$, and \ar, respectively.  
The results are displayed in Fig.~\ref{fig:tnse}, 
where different colors except for gray indicate different known classes. 
Points in gray represent all unknown classes.
The features learned by ERM and One Ring-S cannot be reasonably separated, where the boundaries between known and unknown classes are ambiguous to some extent. 
By contrast, \ar\ provides more meaningful embedding features to distinguish known and unknown samples.




\section{Conclusion}
We investigate the problem of DGCS, 
which is realistic but has been largely overlooked in the literature. 
Specifically, we present a simple yet surprisingly effective approach (\ar) to regularize the model's decision boundary in training and adjust the source-trained classifier's prediction at test time, endowing the deep model with unknown-aware ability even without any access to real data in training. 
Experiments show that \ar\ consistently improves the generalization capability of deep networks in different tasks.
We hope our work will motivate future research on open-world generalization in safety-critical applications.

\section*{Acknowledgement}
This work was partially supported by Hong Kong Research Grants Council under Collaborative Research Fund (Project No. HKU C7004-22G).
It was also partially supported by NSFC62172348, Outstanding Young Fund of Guangdong Province with No. 2023B1515020055, and Shenzhen General Project with No. JCYJ20220530143604010.

{\small
\bibliographystyle{ieee_fullname}
\bibliography{main}
}

\end{document}